\documentclass[letterpaper]{article} 
\usepackage{aaai25}  
\usepackage{times}  
\usepackage{helvet}  
\usepackage{courier}  
\usepackage[hyphens]{url}  
\usepackage{graphicx} 
\usepackage{natbib}  
\usepackage{caption} 
\usepackage{algorithm}
\usepackage{algorithmic}

\usepackage{newfloat}
\usepackage{listings}
\DeclareCaptionStyle{ruled}{labelfont=normalfont,labelsep=colon,strut=off} 
\floatstyle{ruled}
\newfloat{listing}{tb}{lst}{}
\floatname{listing}{Listing}
\usepackage{subfigure}
\usepackage{amsmath}
\usepackage{amssymb}
\usepackage{mathtools}
\usepackage{amsthm}
\usepackage[capitalize,noabbrev]{cleveref}
\theoremstyle{plain}

\theoremstyle{definition}

\theoremstyle{remark}
\usepackage{verbatim}
\usepackage{tikz}

\usepackage[textsize=tiny]{todonotes}
\usepackage{multirow}

\title{Mitigating Feature Gap for Adversarial  Robustness \\ by Feature Disentanglement}
\author{
    Nuoyan Zhou\textsuperscript{\rm 1}\equalcontrib, Dawei Zhou\textsuperscript{\rm 1}\equalcontrib, Decheng Liu\textsuperscript{\rm 1}, Nannan Wang\textsuperscript{\rm 1}\thanks{Corresponding author.}, Xinbo Gao\textsuperscript{\rm 2}
}
\affiliations{
    \textsuperscript{\rm 1}State Key Laboratory of Integrated Services Networks, Xidian University, Xi'an, China\\
    \textsuperscript{\rm 3}Chongqing Key Laboratory of Image Cognition, Chongqing University of Posts and Telecommunications, Chongqing, China\\
    nuoyanzhou@stu.xidian.edu.cn, 
    dwzhou.xidian@gmai1.com, 
    \{dchliu, nnwang\}@xidian.edu.cn, 
    gaoxb@cqupt.edu.cn

}

\usepackage{bibentry}

\begin{document}

\maketitle

\begin{abstract}
 Adversarial fine-tuning methods enhance adversarial robustness via fine-tuning the pre-trained model in an adversarial training manner. However, we identify that some specific latent features of adversarial samples are confused by adversarial perturbation and lead to an unexpectedly increasing gap between features in the last hidden layer of natural and adversarial samples. To address this issue, we propose a disentanglement-based approach to explicitly model and further remove the specific latent features. We introduce a feature disentangler to separate out the specific latent features from the features of the adversarial samples, thereby boosting robustness by eliminating the specific latent features. Besides, we align clean features in the pre-trained model with features of adversarial samples in the fine-tuned model, to benefit from the intrinsic features of natural samples. Empirical evaluations on three benchmark datasets demonstrate that our approach surpasses existing adversarial fine-tuning methods and adversarial training baselines.
\end{abstract}

%

\section{Introduction}
\label{introd}
Deep neural networks (DNNs) have shown impressive performances in various domains of machine learning. However, it has been demonstrated that DNNs are susceptible to adversarial samples and the prediction can be easily manipulated~\citep{FGSM}. Adversarial samples deceive DNNs by introducing imperceptible noise to clean samples. The presence of adversarial samples poses a growing potential threat~\citep{PGD, AA, yu2021lafeat, wei2022physical, wei2023hotcold, liu2024adv, xia2024inspector}, making it crucial to enhance the robustness of networks.

\begin{figure}[!htbp]
\centerline{\includegraphics[width=1.0\columnwidth]{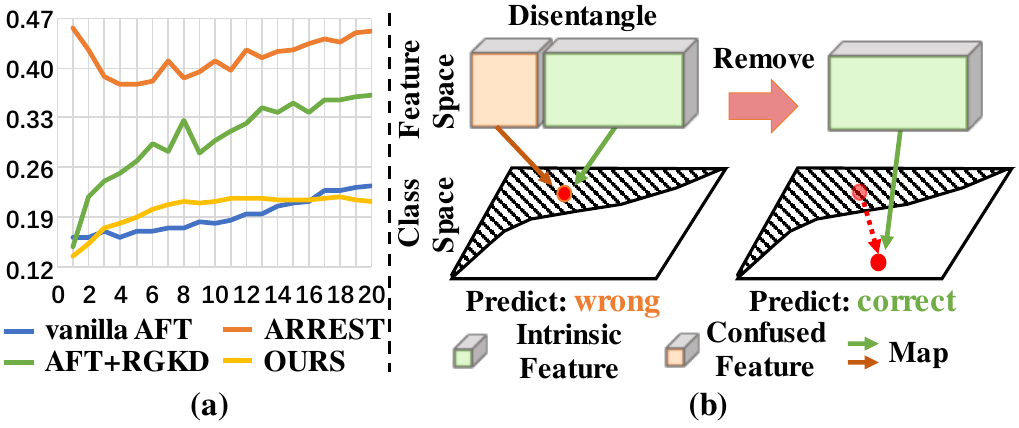}}
\vskip -0.05in
\caption{Illustration of our motivation. (a) $L_{\infty}$ distances of features between natural and adversarial samples during fine-tuning. Previous AFT methods exhibit an increasing trend of the feature gap. (b) Toy illustration of Disentanglement. We model features of adversarial samples and then remove the specific latent features confused by adversarial perturbation to correct the prediction.}
\label{issue_illustration}
\vskip -0.1in
\end{figure}

Many defensive techniques are proposed including adversarial training (AT)~\citep{PGD, TRADES, mart, AWP, Self-AT, TE, LAS-AT, S2O, yu2022understanding, MAN}, adversarial fine-tuning (AFT)~\citep{ARREST, MSD, SAT, multi-robust-lp-convex-hull, lr-schedule, RiFT, si2020better, Agarwal_2023_CVPR, twins}, adversarial purification~\citep{yoon2021adversarial, sun2023critical, lee2023robust, allen2022feature, nie2022diffusion}, adversarial detection~\citep{hickling2023robust, zheng2018robust, pang2018towards}, etc. AT is regarded as the most effective defensive technique among them, but it suffers from several times the training cost as standard (natural) training. To save the cost of training time, AFT employs the same loss function as AT to fine-tune the pre-trained model within a few epochs, which has aroused wide attention.

We identify an issue of the increasing feature gap by analyzing three models fine-tuned by three AFT methods: vanilla AFT~\citep{lr-schedule}, AFT+RGKD, ARREST~\citep{ARREST}. As illustrated in Figure~\ref{issue_illustration} (a), the feature distance between natural and adversarial samples abnormally maintains an increasing direction during the fine-tuning process. Though ARREST initially exhibits notable fluctuations with a decreasing trend, the overall magnitudes of feature distance in three AFT methods show an increasing trend. We suppose that some specific latent features are easily confused by adversarial attacks and lead to the growing gap between the features of natural and adversarial samples. Considering that a robust model should treat natural and adversarial samples equally and extract similar features from them, we expect to remove the specific latent features causing the feature gap to achieve better robust performance, as shown in Figure~\ref{issue_illustration} (b).

In this paper, we analyze the specific latent features causing the feature gap and introduce an approach based on disentanglement to explicitly model and remove these features. We propose a new technique named \textbf{A}dversarial \textbf{F}ine-tuning via \textbf{D}isentanglement (\textbf{AFD}). First, we disentangle the features of adversarial samples as two components, in which the specific latent features causing the feature gap are identified as the features confused by adversarial perturbation. Next, we propose a feature disentangler to model and separate them from the features of adversarial samples. We maximize the predicted probabilities of the wrong predicted classes to acquire the specific latent features. Then we impose a constraint to distance the features of adversarial samples from them, to eliminate the specific latent features and mitigate the feature gap. Besides, we align features of adversarial samples in the fine-tuned model with features of natural samples in the naturally pre-trained model, further eliminating the specific latent features. Experiments demonstrate that our AFD alleviates the feature gap and improves the robustness. Contributions are summarized as follows: 

\begin{itemize}
\item We observe the gap in features between natural and adversarial samples anomalously increases in adversarial fine-tuning methods, resulting from the specific latent features confused by adversarial perturbation. 
\item We provide the theoretical analysis of feature disentanglement. Then we introduce a disentanglement-based AFT approach, which explicitly models and further eliminates the specific latent features causing the feature gap by disentanglement and alignment. 
\item Empirical evaluations show our approach mitigates the gap in features between natural and adversarial samples and surpasses existing methods. We also provide extended analyses for a holistic understanding.
\end{itemize}

\section{Related Work}
\label{related}
\textbf{Adversarial Attack} Adversarial Attack is first introduced in \citet{FGSM}, which fools DNNs by imperceptible perturbations. \citet{PGD} propose Projected Gradient Descent (PGD) attack based on the gradient projection and random start. \citet{AA} introduce a powerful adaptive attack (AutoAttack), including three white-box attacks and a black-box attack. \citet{yu2021lafeat} demonstrate that exploiting latent features is highly effective against many defense techniques. Besides, adversarial attacks can also work in the physical world.~\citet{wei2022physical} present a comprehensive overview of physical adversarial attacks.~\citet{wei2023hotcold} propose a novel physical adversarial attack that applies the Warming Paste and Cooling Paste to hide persons from being detected.~\citet{liu2024adv} learn adversarial semantic perturbations in the latent space for high attack capabilities and low perceptibility. We take PGD and AutoAttack to evaluate robustness.

\textbf{Adversarial Training} Adversarial training (AT) is the most effective technique to defend against attacks. \citet{PGD} propose PGD-based adversarial training (PGD-AT), compelling the model to correctly classify adversarial samples within the epsilon sphere during training. \citet{TRADES} reduce the divergence of probability distributions of natural and adversarial samples to mitigate the difference between robust and natural accuracy. \citet{xie2019feature} apply non-local means or other filters as the denoising block. \citet{mart} find that misclassified samples harm adversarial robustness significantly, and propose to improve the model's attention to misclassification by adaptive weights. \citet{yan2021cifs} manipulates channels’ activations to align the channel with the relevance to predictions. \citet{yang2021adversarial} introduce a disentanglement-based architecture to generate class-specific and class-irrelevant representations. \citet{MAN} embed a label transition matrix into models to infer natural labels from adversarial noise. \citet{kim2023feature} propose the FSR module to separate non-robust activations and recalibrate the features. \citet{zhou2023enhancing} define two attributes of robust features to guide robust training. Though AT has good robust performances, it suffers from a large quantity of computing expenses. Some work has been developed to accelerate adversarial training, commonly known as fast adversarial training (FAT)~\citep{Wong2020FastIB, Kim2020UnderstandingCO, huang2023fast}. Since these FAT methods only require a little training time like AFT, we compare them with our method.

\textbf{Adversarial Fine-tuning} Adversarial fine-tuning (AFT) has been employed to enhance the pre-trained model within a few epochs for various targets, particularly for achieving adversarial robustness. \citet{MoosaviDezfooli2018RobustnessVC} analyze the effect of AFT by comparing the decision boundary of a DNN before and after applying AFT. \citet{kumari2019harnessing, yu2021lafeat} discover the vulnerability of latent features in adversarial trained models and propose adversarial approaches to mitigate this. \citet{lr-schedule} employ vanilla AFT to expedite the training of robust models with a few training epochs and a warm-up learning rate schedule. Some researchers~\citep{MAX/AVG,MSD,SAT,multi-robust-lp-convex-hull} aim to achieve multiple $l_p$ adversarial robustness through specific strategies. Based on representation learning,~\citet{ARREST} constrain the representation to mitigate the accuracy-robustness trade-off. \textbf{Although they aim to learn suitable features, the gap between natural and adversarial features can't be eliminated by simple feature constraint, which has been indicated in Figure~\ref{issue_illustration} (a)}.~\citet{RiFT} propose a metric to measure the robustness of modules and fine-tune them for better out-of-distribution performance. \citet{twins} introduce a two-pipeline structure to improve the relationship between the weight norm and its gradient norm in batch normalization layers. Despite the extensive work, few methods (vanilla AFT~\citep{lr-schedule}, ARREST and AFT+RGKD~\citep{ARREST}) are based on naturally pre-trained models, which are our important baselines.

\begin{figure*}[t]
\begin{center}
\centerline{\includegraphics[width=2.0\columnwidth]{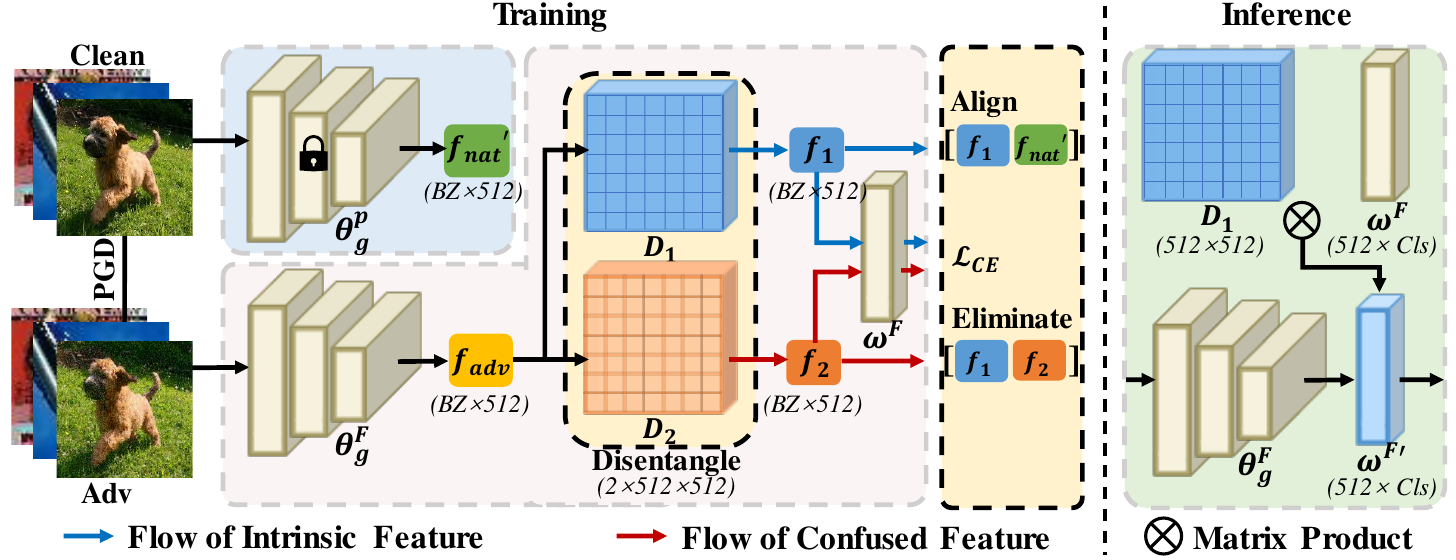}}
\caption{Overview of our AFD. Lock denotes the frozen parameters, $BZ$ denotes the batch size, $Cls$ denotes the class number, and numbers in brackets denote the sizes of features or parameters. During fine-tuning, Adversarial features are disentangled into $f_{1}$ and $f_{2}$. $\mathcal{L}_{CE}$ ensures $f_{2}$ to approximate confused features. We keep $f_{1}$ away from $f_{2}$ to eliminate confused features in $f_{1}$. Besides, we align $f_{1}$ with pre-trained intrinsic features $f_{nat}^{\prime}$ to further correct the prediction. During inference, we multiply $D_{1}$ and ${\omega}^F$ to obtain the robust model without additional modules.
}
\label{structure}
\end{center}
\vskip -0.1in
\end{figure*}

\section{Preliminaries}
\label{preliminarie}
We denote a DNN as $h(x) = \omega \cdot g(x; \theta_g)$, where $x$ is the input image, and $\theta_g$ is the parameters of the feature extractor $g$, i.e., the model before the last linear classifier. $\omega$ denotes the parameters of the last linear classifier. The parameters of the naturally pre-trained and fine-tuned model are denoted as $\theta^{P}$ and $\theta^{F}$, respectively. The model parameters $\theta^{F}$ are initialized by $\theta^{P}$. $x$ denotes the clean sample, and $x^{\prime}$ denotes the adversarial sample. $\mathbb{N}(x, \epsilon)$ represents $\{x^{\prime}:\|x^{\prime}-x\|_{\infty} \leq \epsilon\}$, $\epsilon$ is the perturbation budget, ${\Vert{\cdot}\Vert}_p$ denotes the $l_p$ norm. Adversarial and natural features denote the features in the last hidden layer of adversarial and natural samples, i.e., $g(x^{\prime}; \theta_g)$ and $g(x; \theta_g)$ respectively.

\section{Methodology}
In this section, we first reveal the issue of the increasing gap of features between natural and adversarial samples in AFT. Then we introduce a disentanglement-based method, to model and further remove the specific latent features leading to the feature gap. Our method is expected to mitigate the feature gap to enhance adversarial robustness.

\subsection{Specific Latent Feature Causing Feature Gap}
\label{MSL}

As shown in Figure~\ref{issue_illustration} (a), there is a growing gap in features between natural and adversarial samples in existing AFT methods. Intuitively, the ideal robust model should have a consistent understanding and analysis of natural and adversarial samples, thus extracting similar (the same) features from them. Thus, narrowing the feature gap is expected to achieve better robust performance.

To realize the target, we can disentangle two types of features from the features of natural and adversarial samples. We define natural features $f_{nat}$ as the intrinsic features $f_{i}$.  We can define adversarial features $f_{adv}$ as the combination of the intrinsic features $f_{i}$ and the confused features $f_{c}$, the latter of which leads to the gap in features between natural and adversarial samples. In this modeling, the intrinsic features represent the features that contribute to accurate classification, which are easily extracted from natural samples. Besides, the confused features are identified as the specific latent features causing the gap between features of natural and adversarial samples, which are confused by the adversarial perturbation. As illustrated in Figure~\ref{issue_illustration} (b), if the confused features are removed in adversarial features, the features of adversarial samples can align with those of natural samples, resulting in a zero feature gap and accurate predictions for adversarial samples.

We suppose the phenomenon in Figure~\ref{issue_illustration} (a) is attributable to the presence of the confused features $f_{c}$. Because the confused features persist as latent features and don't diminish during fine-tuning, the gap between features of natural and adversarial samples cannot be bridged. The confused features induce the gap between natural and adversarial features, leading to limited robust performance. Based on our observations and analyses, we introduce a new approach named \textbf{A}dversarial \textbf{F}ine-tuning via \textbf{D}isentanglement (\textbf{AFD}), as shown in Figure~\ref{structure}. It applies a feature disentangler to explicitly extract the confused features, and remove them by the elimination and alignment of features.

\subsection{Adversarial Fine-tuning Based on Disentanglement}
\label{dis}

The primary strategy is to explicitly disentangle the intrinsic and confused features from the features of adversarial samples. We propose a feature disentangler to model confused features for further elimination. The features in the last hidden layer are input into the disentangler, yielding two disentangled feature vectors. We define these feature vectors as intrinsic features $f_{1}$ and confused features $f_{2}$. To implement the disentanglement, we introduce a disentangler with two linear blocks $D_1, D_2$ between the feature extractor $g$ and the linear classifier $\omega$. These linear blocks are expected to disentangle adversarial features into intrinsic and confused features. The disentangled feature vectors $f_{1}$ and $f_{2}$ are constrained to approximate the intrinsic and confused features. Given by: 
\begin{equation}
\begin{split}
&f_{1} = {D_1}(g(x^{\prime}; \theta_g^F)),  \\
&f_{2} = {D_2}(g(x^{\prime}; \theta_g^F)),  \\
&{\rm where}  \ \left\{
             \begin{array}{lr}
             f_{1} \approx f_{i}, & \\
             f_{2} \approx f_{c}, & 
             \end{array}
\right.
\label{eq_target}
\end{split}
\end{equation}
where $\theta_g^F$ denotes the parameters of the feature extractor in the fine-tuned model. $x^{\prime}$ denotes the adversarial samples created by PGD-10~\citep{PGD}, formulated as:

\begin{equation}
x^{\prime} = \mathop{\boldsymbol{\Pi}} \limits_{ \mathbb{N}(x, \epsilon)}\left(x + \epsilon \cdot \operatorname{sign}\left(\nabla_{x} \mathcal{L}_{CE}\left(\omega^F \cdot{D_1}(g(x; \theta_g^F)), y\right)\right)\right),
\label{adv}
\end{equation}

where $\boldsymbol{\Pi}$ denotes a clamp function, $\operatorname{sign}$ denotes a sign function, $\mathcal{L}_{CE}$ denotes the cross-entropy loss. We take $f_{1}$ and $f_{2}$ to match the intrinsic and confused features. The input and output dimensions of $D_1, D_2$ are set to match the channel dimensions of the feature vectors before disentanglement. This ensures the parameters of the linear block $\theta_{D_1}$ can be seamlessly integrated with the parameter of the last linear classifier $\omega^F$, expressed as $\theta_{D_1} \cdot \omega^{F} = {\omega^{F\prime}}$. Therefore, we only need to load the linear classifier $\omega^{F\prime}$ without additional disentangling modules in the inference (test) phase.

To achieve good performance on both accuracy and robustness, we should design suitable loss functions to realize Formula~\ref{eq_target}. We suggest fitting $f_{2}$ to incorrectly predicted labels $y^{\prime}$ to model the confused features. The confused features result in obvious shifts in both features and predicted probabilities, leading to wrong predicted class labels in most cases. Thus, it can be identified as the specific latent features causing the gap between natural and adversarial features. To decrease the dependence between the confused features and ground-truth labels, we force the disentangled features $f_{2}$ to match incorrect class labels, formulated as:
\begin{equation}
\begin{split}
\mathcal{L}_{1} =\ & \mathcal{L}_{CE}(\omega^F \cdot f_{2}, y^{\prime}), \\
 {\rm where} \quad &y^{\prime} = {\rm \mathop{max}\limits_{y^{\prime} \neq y}}\ (\omega^F \cdot f_{1}) ,
\end{split}
\label{eq_1}
\end{equation}
where $y^{\prime}$ denotes wrong predicted labels with the maximum predicted probability. We only optimize the disentangling block $D_2$ instead of the entire model. This selective optimization prevents potential harm to the feature extractor that could result from learning the confused features.

In addition, we introduce a constraint on $f_{1}$ to keep away from the confused features $f_{2}$, removing the confused features from the features of adversarial samples. The exclusive force between $f_{1}$ and $f_{2}$ encourages the elimination of the confused features from the features of adversarial samples. This constraint aims to protect adversarial features from being perturbed by the confused features, and reduces the association between confused and intrinsic features. Given by:
\begin{equation}
\mathcal{L}_{2} = \ -\mathcal{D}(f_{1}, f^{-}_{2}), 
\label{eq_2}
\end{equation}
where $\mathcal{D}$ denotes a distance function, $f^{-}_{2}$ means the gradient back-propagation of $f_{2}$ is frozen.

\subsection{Adversarial Fine-tuning Based on Alignment}
\label{ali}
The second strategy is to align the adversarial features in the fine-tuned model with the natural features in the pre-trained model. Since the pre-trained model has a high natural accuracy and doesn't suffer from adversarial attacks, the natural features in the pre-trained model can be identified as ideal intrinsic features. Thus, adversarial features in the fine-tuned model can be greatly improved by being aligned with the natural features in the pre-trained model. Besides, lots of work~\citep{TRADES, mart, ARREST} conducts an alignment between natural and adversarial features, which is regarded to mitigate the trade-off between accuracy and robustness. We take the disentangled features $f_{1}$ to match the natural features in the pre-trained model, further removing the confused features from $f_{1}$. The AFT loss function with the alignment can be formulated as:
\begin{equation}
\mathcal{L}_{3} = \mathcal{L}_{CE}(\omega^{F} \cdot f_{1}, y) + \gamma \cdot \mathcal{D}(f_{1}, g(x; \theta_g^P)^{-}),
\label{eq_3}
\end{equation}
where $\theta_g^P$ denotes parameters of the feature extractor in the pre-trained model, $\gamma$ denotes the weight of the alignment, and $g(x; \theta_g^P)^{-}$ denotes the gradient back-propagation of $g(x; \theta_g^P)$ is frozen. Therefore, the objective function of our AFD is defined as follows: 
\begin{equation}
\mathcal{L}_{total} = \alpha \cdot \mathcal{L}_{1} + \beta \cdot \mathcal{L}_{2} + \mathcal{L}_{3}, 
\label{eq_total}
\end{equation}
where $\alpha$ and $\beta$ denote the weights of $\mathcal{L}_{1}$ and $\mathcal{L}_{2}$. The algorithm of AFD in Algorithm~\ref{alg}. 


\begin{algorithm}
\caption{Adversarial Fine-tuning via Disentanglement}\label{alg}
\begin{algorithmic}
\REQUIRE Training set $\mathcal{D}$, Hyperparameters $\alpha, \beta$, Number of epochs $T$, Learning rate $\eta$, Naturally pre-trained model parameters $\theta^P$, Model parameters except linear block $D_2$ parameters $\theta_1$, Linear block $D_2$ parameters $\theta_2$.
\ENSURE Fine-tuned model parameters $\theta_1$.
\STATE Initialize $\theta_1$ by $\theta^P$;
\FOR{$t=1, \cdots, T$}
    \FOR{$x, y \in \mathcal{D}$}
        \STATE Calculate $x^{\prime}$ by by Formula~\ref{adv};
        \STATE Calculate features $f_1, f_2$ by Formula~\ref{eq_target};
        \STATE Calculate loss $\mathcal{L}_{1}$ Formula~\ref{eq_1};
        \STATE $\theta_{2} = \theta_{2} - \eta \cdot \frac{\mathrm{d} (\alpha\cdot\mathcal{L}_{1})}{\mathrm{d} \theta_{2}}$;
        \STATE Calculate loss $\mathcal{L}_{2}$ Formula~\ref{eq_2};
        \STATE $\theta_{1} = \theta_{1} - \eta \cdot \frac{\mathrm{d} (\beta\cdot\mathcal{L}_{2})}{\mathrm{d} \theta_{1}}$;
        \STATE Calculate loss $\mathcal{L}_{3}$ by Formula~\ref{eq_3};
        \STATE $\theta_{1} = \theta_{1} - \eta \cdot \frac{\mathrm{d} \mathcal{L}_{3}}{\mathrm{d} \theta_{1}}$;
    \ENDFOR
\ENDFOR
\end{algorithmic}
\end{algorithm}

\section{Experiments}
\label{experiment}
This section presents experiments with AFD. We first introduce our experiment setting. We further make comparisons on multiple architectures and datasets to show the effectiveness. Then we conduct an ablation study to show the functions of the disentangling module and alignment. Finally, we conduct an empirical analysis to support our hypothesis. 

\subsection{Setting}

\textbf{Datasets}  We conduct experiments on three benchmark datasets including CIFAR-10 and CIFAR-100~\cite{CIFAR-10}, Tiny-ImageNet~\cite{TinyImageNet}. CIFAR-10 dataset contains 60,000 color images having a size of $32\times32$ in 10 classes, with 50,000 training and 10,000 test images. CIFAR-100 dataset contains 50,000 training and 10,000 test images in 100 classes. Tiny-ImageNet dataset contains 100000 images of 200 classes (500 for each class) downsized to 64×64 colored images. Each class has 500 training images, 50 validation images and 50 test images. We only use training and validation images of Tiny-ImageNet. 

\textbf{Baselines} To make a comprehensive comparison, we have employed various techniques to demonstrate the effectiveness of our method AFD. Our primary baselines are AFT methods including vanilla AFT (vAFT)~\citep{lr-schedule}, AFT+RGKD (AFKD)~\citep{ARREST}, and ARREST~\citep{ARREST}. All of them are built upon naturally pre-trained models. Additionally, we compare our AFD with a state-of-the-art AFT method RiFT~\citep{RiFT}, which relies on adversarially pre-trained and has powerful out-of-distribution performances. Besides, we select several advanced fast adversarial training methods, including FreeAT~\citep{Shafahi2019AdversarialTF}, FGSM-GA~\citep{Andriushchenko2020UnderstandingAI}, ATAS~\citep{huang2023fast}, since they are also low-cost techniques. Moreover, we compare with baseline methods of adversarial training including PGD-AT~\citep{PGD}, TRADES~\citep{TRADES} to show our great advance in robustness.

\textbf{Optimization Details}
The optimizer is Stochastic Gradient Descent (SGD) optimizer with a momentum of 0.9. We fine-tune the pre-trained models with a batch size of 128, a learning rate of 0.0025, a weight decay of ${5.0\times10^{-4}}$, and training epochs of 20. The schedule of learning rate is the same as~\citet{ARREST}. We utilize Exponential Moving Average (EMA)~\citep{EWA} to gain better parameters.

\textbf{Implementation Details}
 We use ResNet18~\cite{ResNet} and WideResNet-34-10~\cite{WideResNet} as the main DNN architectures, following previous studies~\citep{ARDgoldblum2020adversarially, ARREST, mart}. The distance function chosen for $\mathcal{L}_{2}$ and $\mathcal{L}_{3}$ is the angular distance: $\mathcal{D}(u, v) = 1 - \frac{u\cdot v}{{\Vert{u}\Vert}_2\cdot{\Vert{v}\Vert}_2}$. For hyperparameters $\{\alpha, \beta, \gamma\}$, we suggest $\{0.05, 0.25, 25\}$ in most cases.  We report the natural accuracy (Clean), the robust accuracy against PGD-20 (PGD)~\citep{PGD}, the robust accuracy against AutoAttack (AA)~\citep{AA}, and the average value (Avg) for evaluation. Since $D_1$ is seamlessly integrated with ${\omega}^F$ at inference time, we don't design adaptive attacks about $f_1$. The maximum perturbation is set to $8/255$. $L_{\infty}$-norm PGD with a random start, a step size of 0.003, and attack iterations of 20 is utilized in the evaluation. In the following tables, $\dag$ denotes the results excerpted in the paper~\citep{huang2023fast}, which are evaluated by PGD-10 in the manuscript. \pmb{Bold} and \underline{underline} indicate the highest and second-highest values for each metric. We only report \textbf{the results of the last epoch}, since AFT methods always gain the best performance in the last epoch.

\begin{table*}[h]

\begin{center}
\begin{tabular}{c|cccc|cccc|cccc}
\hline

\multicolumn{1}{c|}{\multirow{2}*{Methods}}& \multicolumn{4}{c|}{CIFAR-10}& \multicolumn{4}{c|}{CIFAR-100} & \multicolumn{4}{c}{Tiny-ImageNet} \\
\cline{2-5} \cline{6-9} \cline{10-13}
\multicolumn{1}{c|}{}&Clean&PGD&AA&Avg&Clean&PGD&AA&Avg&Clean&PGD&AA&Avg\\
\hline

PGD-AT    &   84.43&   48.82 &   43.68&  58.98&   59.09&   24.05 & 20.67 & 34.60 
 & 52.12  &  19.38 &  16.79 &   29.43 \\
TRADES &  82.37&  \pmb{53.27}&  \pmb{48.52} & \underline{61.39} & 55.47 &  \underline{28.12} & \pmb{23.49}& 35.69 &   49.28   &  \underline{22.59}&  \underline{17.06} &    29.64 \\
\hline
Free-AT$^{\dag}$    &   78.37&   40.90 &   36.00& 51.76&   50.56&   19.57 & 15.09 & 28.41&-&-&-&-\\
FGSM-GA$^{\dag}$  &  80.10&  49.14&  43.44 & 57.56 &50.61 & 24.48&  19.42 &31.50&-&-&-&-\\
ATAS$^{\dag}$  &  81.22&  50.03&  45.38 & 58.88& 55.49 & 27.68&  22.62& 35.26 &-&-&-&-\\
\hline
vAFT & 83.93  &  51.40  & 45.48 &60.27 & 60.83 & 26.55  &  21.96 & 36.45 & 52.75  &  21.29  & 16.76 &    30.27   \\
RiFT     & 84.49  &  49.24  & 43.66  & 59.13 &   59.51&   24.01 & 20.48 & 34.67 & 52.24  &  20.89  & 17.05  &    30.06  \\
AFKD    & 84.71&  51.32 &   46.09&  60.71 &  65.54&  28.09 & 22.92 & \underline{38.85} & 57.17&  22.29 &  16.68 &    \underline{32.05}     \\
ARREST &  \pmb{85.71}&  51.23&  46.23 &  61.05 & \pmb{67.00} & 26.89&  21.34 & 38.41 & \pmb{60.36} & 19.28 & 12.58 &   30.74    \\
AFD &  \underline{84.88}&  \underline{52.70} &  \underline{47.40} & \pmb{61.66} & \underline{65.69} & \pmb{28.56} &  \underline{23.05} & \pmb{39.10} &  \underline{57.41} & \pmb{23.63} & \pmb{17.73}&  \pmb{32.92}  \\ 
\hline

\end{tabular}
\end{center}
\caption{Quantitative evaluations of all methods on ResNet18.}
\label{res18}
\end{table*}

\begin{table*}[h]

\begin{center}
\begin{tabular}{c|cccc|cccc|cccc}
\hline
\multicolumn{1}{c|}{\multirow{2}*{Methods}}& \multicolumn{4}{c|}{CIFAR-10}& \multicolumn{4}{c|}{CIFAR-100}& \multicolumn{4}{c}{Tiny-ImageNet}  \\
\cline{2-5} \cline{6-9} \cline{10-13}
\multicolumn{1}{c|}{}&Clean&PGD&AA&Avg&Clean&PGD&AA&Avg&Clean&PGD&AA&Avg\\
\hline
PGD-AT & 86.35& 49.57 & 45.81 &     60.58    & 60.30 &  25.14  & 22.65 & 36.03  &  54.24  &  19.03 & 16.51&   29.93 \\
TRADES & 85.68 &  52.53  &  49.22 &   62.48      &57.67 &  28.26  & 25.25 & 37.06  & 49.22 & 23.33&  18.51&  30.35  \\ 
\hline

vAFT  & 87.42  & 53.57 & 48.49 &   63.16      & 64.92 &  27.19 &  23.59 & 38.57  &  57.17  &   22.24    &  18.02&   32.48  \\
RiFT     & 84.49  &  49.24  & 45.83  &   59.85      & 60.59   &   25.57    &   22.67 & 36.28 &55.09  &   19.51    &  16.79&  30.46   \\
AFKD    & \pmb{88.54}&  \underline{54.86} &  \underline{49.57} &    \underline{64.31}   &68.50  &  \underline{30.12} &  \underline{25.28} & \underline{41.30}  &60.92  & \underline{24.27}  &  \underline{19.01} &  34.73  \\
ARREST &  \underline{88.42}&  54.16&  49.45 &   64.01       & \pmb{69.52} & 29.47 & 24.55 & 41.18 &\pmb{65.07} & 22.73 & 16.67& \underline{34.82} \\
AFD &  88.08& \pmb{56.12}&  \pmb{51.62} &  \pmb{65.27}   & \underline{68.51}&\pmb{32.63}&\pmb{27.62} & \pmb{42.92} &  \underline{63.89}  & \pmb{25.73} & \pmb{19.13} &  \pmb{36.25} \\
\hline

\end{tabular}
\end{center}
\caption{Quantitative evaluations of all methods on WideResNet-34-10.}
\label{wide}
\end{table*}

\subsection{Main Results}

\textbf{ResNet18} First, AFD effectively mitigates the trade-off between generalization and robustness. As shown in Table~\ref{res18}, AFD takes the lead in robust performance and exhibits the second-best clean accuracy \textbf{among AFT methods}. It gains less clean accuracies compared to ARREST by 1.70\%, but its robust accuracies against PGD-20 and AA outperform ARREST by 2.50\% and 2.68\%. The improvement in robustness is evidently larger than the drop in clean accuracy. This result indicates the positive effect of feature disentanglement on the accuracy-robustness trade-off. 

Moreover, AFD exceeds almost all the AT and FAT methods. Although the powerful AT baseline TRADES outperforms AFD in the robust accuracies on CIFAR-10 and CIFAR-100 by 0.56\%, it surpasses TRADES in clean and average accuracy by 6.365\% and 1.84\%. Besides, it surpasses TRADES on Tiny-ImageNet in all metrics. The result implies that efficient AFD can achieve comparable performances with AT.

\textbf{WideResNet-34-10} Second, AFD takes further advantages on large architectures. As shown in Table~\ref{wide}, AFD maintains the best robust and average accuracies among all the approaches. It exceeds the second-best baselines by approximately 1.62\% and 1.34\%. The enhancement of robustness further enlarges compared to that on ResNet18. In addition, the natural generalization of AFD always ranks second and is less than the best clean accuracy by 0.88\% on average. The discrepancy of clean accuracies between AFD and the best one is smaller than that on ResNet18 (1.69\%). Furthermore, AFD outperforms AT baselines in all metrics. These demonstrate that AFD achieves a better trade-off between generalization and robustness on larger architectures.

It's worth noting that AFD doesn't exhibit optimal natural generalization among AFT methods. It is attritued to merely adversarial features purified by disentangling modules. Intuitively, natural features can also be boosted by disentanglement mechanism. Nevertheless, it will result in additional computing cost. Besides, current AFD has remained the second-best clean accuracy among various methods.

\textbf{Summary} These experiments indicate that AFD always outperforms all the baselines for different attacks, datasets, and architectures in the aspect of overall performance. It proves the effectiveness of our AFD.

\subsection{Ablation Study}
\label{Ab}

\textbf{Disentangler and Alignment} As shown in Table~\ref{ablation}, vanilla AFT with the disentangler (vAFT+D) exhibits superior performances compared to vanilla AFT by 0.70\%, 1.28\%, and 1.40\%, respectively. It shows that the disentangler effectively eliminates the confused features that lead to wrong predictions. Besides, vanilla AFT with the alignment of features (vAFT+A) outperforms vanilla AFT by 0.68\%, 0.31\%, and 1.33\%. It indicates that the alignment successfully contributes to removing the confused features. Moreover, the combination of disentangler and alignment (AFD) has the best performance on both natural and robust accuracy, which shows two strategies are compatible and the combination can boost the positive effects of individual strategies.

\begin{table}[!htbp]

\begin{center}
\begin{tabular}{cccccc}
\hline
\multicolumn{1}{c}{Methods}&vAFT&vAFT+D&vAFT+A&AFD\\
\hline
Clean & 83.93  &  84.63  & 84.61& 84.88\\
PGD & 51.40  & 52.68  & 51.71 & 52.70  \\
AA & 45.48&  46.88 &   46.81 &  47.40 \\
\hline
\end{tabular}
\end{center}
\caption{The performances of vAFT and other methods on CIFAR-10 on ResNet18. `+D' denotes the method with the disentangler, and `+A' denotes the method with the alignment of features.}
\label{ablation}
\vskip -0.1in
\end{table}

\textbf{Hyperparameter} There are three hyperparameters in our AFD: $\alpha, \beta$ and $\gamma$. As illustrated in Figure~\ref{hyper}, there is a remarkable trade-off between accuracy and robustness when $\beta$ and $\gamma$ are growing. On the one hand, a larger value of $\beta$ indicates a larger exclusive force between the intrinsic and confused features, leading to less non-robust features utilized in classification and better robustness. On the other hand, as $\gamma$ increases, the natural accuracy increases but robust accuracy drops. This occurs because more feature information of natural samples is transferred to the fine-tuned model through the feature alignment. Besides, the variations of both natural and robust accuracies are less than 0.60\%, demonstrating that our \textbf{AFD is insensitive to variations of hyperparameters}. This conclusion is also true among different datasets.


\begin{figure}[!htbp]
\begin{center}
\centerline{\includegraphics[width=1.0\columnwidth]{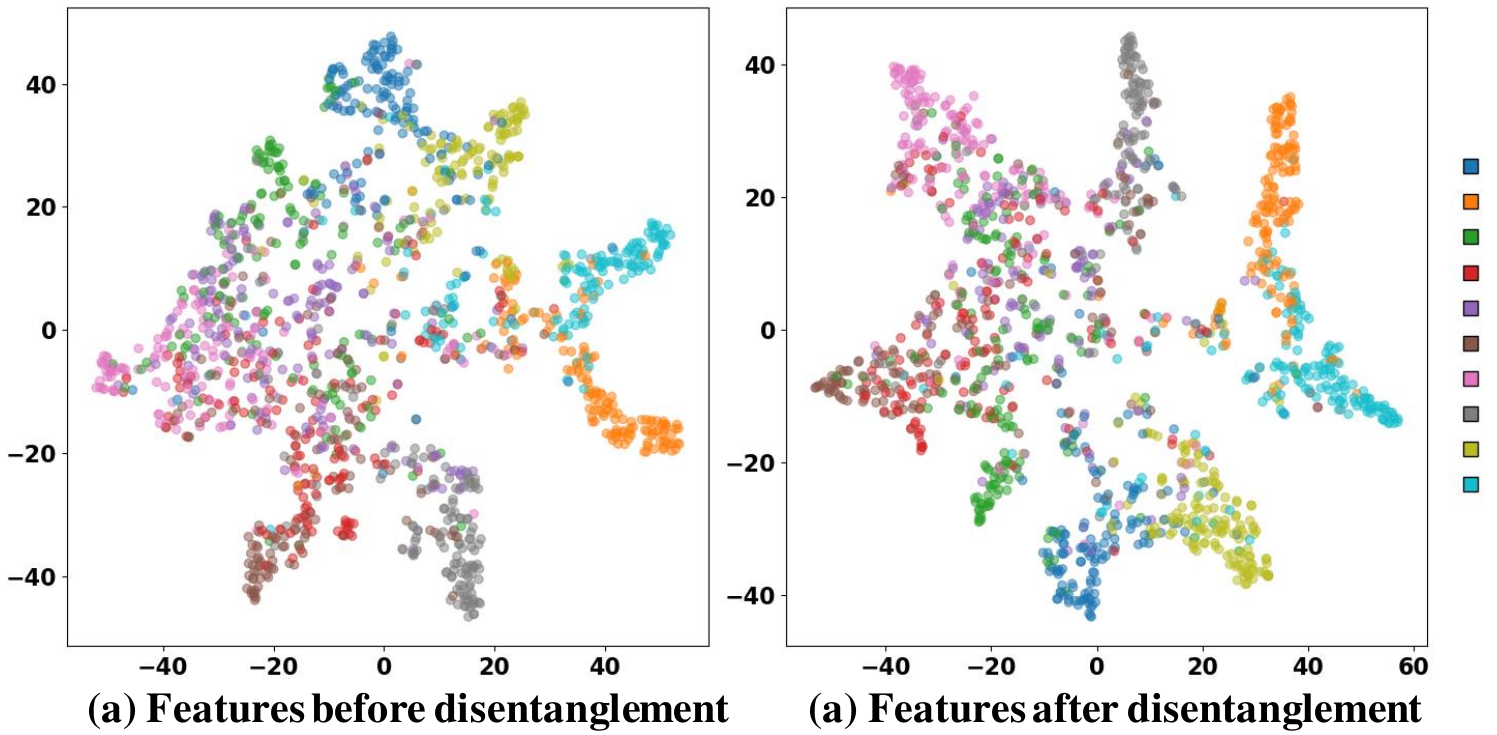}}
\caption{t-SNE visualization on CIFAR-10 on ResNet18.}
\label{dis_compare}
\end{center}
\vskip -0.2in
\end{figure}

\begin{figure*}[t]
\begin{center}
\centerline{\includegraphics[width=2.0\columnwidth]{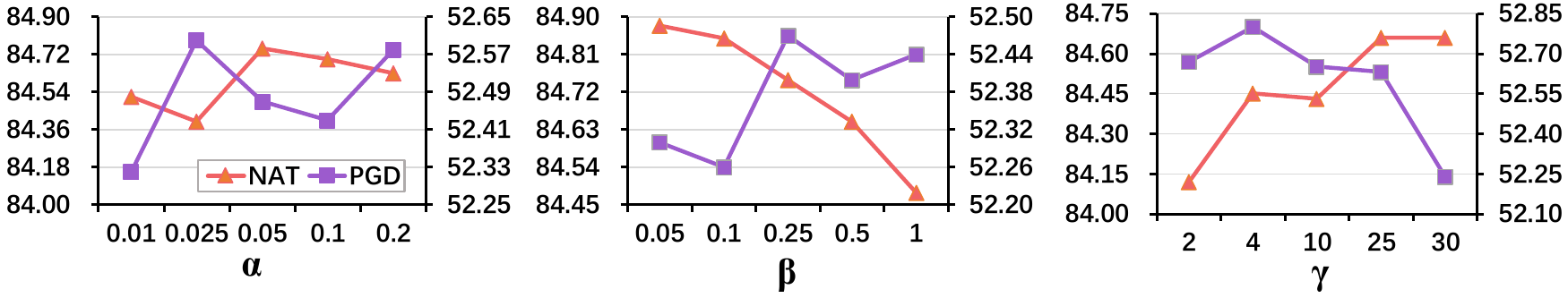}}
\vskip -0.05in
\caption{The clean accuracy (NAT) and robust accuracy against PGD-20 (PGD) of AFD with different hyperparameters.}
\label{hyper}
\end{center}
\vskip -0.1in
\end{figure*}

\subsection{Empirical Analysis of Disentangled Features}

we analyze the feature distance between the intrinsic features $f_{1}$ and natural features $f_{nat}$, as shown in Figure~\ref{issue_illustration} (a). In contrast to the growing trend of feature distance in existing AFT methods, the optimization trajectory of the feature distance in AFD exhibits a fast convergence trend. It implies that AFD has mitigated the gap between natural and adversarial features. Additionally, the magnitudes of the feature distance in AFD are smaller than other AFT methods, e.g., the $L_{\infty}$ distance of AFD (0.212) is notably less than those of ARREST and vanilla AFT (0.453, 0.236). These results demonstrate our AFD can enhance the adversarial robustness by alleviating the feature gap.

Moreover, we evaluate the performances of two types of disentangled features $f_{1}$, $f_{2}$ on various datasets and models. As shown in Table~\ref{f_non}, the natural accuracies of $f_{2}$ are lower than those of $f_{1}$ by $5.67\% \sim 51.04\%$. It indicates the disentangler has successfully extracted and removed the confused features as $f_{2}$. High accuracies of $f_{1}$ indicate models benefit from eliminating the confused features. Interestingly, the disparity of accuracies on CIFAR-100 is smaller than those on CIFAR-10 and Tiny-ImageNet. We suppose that it is challenging to learn clear feature space from a small-resolution dataset with numerous classes, which made the linear disentangler difficult to extract confused features.


We make further feature visualization by t-SNE~\citep{tsne}. We take the fine-tuned model by AFD. The we acquire features of adversarial samples crafted by PGD-20. We visualizes two types of features: the features before disentanglement and the intrinsic features $f_{1}$. As illustrated in Figure~\ref{dis_compare}, features after disentanglement has a distincter class boundary than features before disentanglement, indicating the effectiveness of disentanglement.


\begin{table}[htbp]

\begin{center}
\begin{tabular}{ccccc}
\hline
Model&Data&$f_{1}$&$f_{2}$&$\Delta$\\
\hline
RN18 & CIFAR-10 &  84.88 & 39.05   & 45.83 \\
RN18 & CIFAR-100 & 65.69  & 58.17  & 7.52   \\
RN18 & Tiny-ImageNet & 57.41  & 48.78  &  8.63  \\
\hline
RN50 & CIFAR-10 &  87.01  &  35.97 &  51.04 \\
RN50 & CIFAR-100 & 67.45  & 56.17  &  11.28 \\
\hline
WRN-34-10 & CIFAR-10 &  88.08 & 46.89&   41.19\\
WRN-34-10 &  CIFAR-100 &  68.51  & 62.84 &  5.67 \\
WRN-34-10 & Tiny-ImageNet & 63.89  & 55.15  & 8.74   \\
\hline
\end{tabular}
\end{center}
\caption{The natural accuracies of the intrinsic features $f_{1}$ and confused features $f_{2}$. RN and WRN are abbreviations of ResNet and WideResNet, respectively. $\Delta$ denotes the accuracy disparity.}
\label{f_non}
\end{table}

\section{Training Time and Model Parameter}
\label{env}

We also measure the training time as an evaluation indicator. Notice that the training time of PGD-AT with $\dag$ is almost the same as ours. Thus it is fair to compare our training time with theirs. The vanilla ResNet18 and WideResNet-34-10 have parameters of 11.2M and 46.2M. As shown in Table~\ref{time}, our AFD costs several times less time than AT methods, and its total training time surpasses other AFT approaches merely by less than 200 seconds. Considering its advanced performance, the time cost is worthy. 

 The vanilla ResNet18 and WideResNet-34-10 have parameters of 11.2M and 46.2M. In the training process, the additional module has a weight matrix that doesn’t depend on the input size. It is $2*512*512=0.5$M (4.46\% increase) on ResNet18 and $2*1024*1024=2$M (4.33\% increase) on WideResNet-34-10, which is acceptable for better training.

\begin{table}[!htbp]

\begin{center}
\begin{tabular}{cccc}
\hline
Method&Epoch&${\rm Time_{one}}$& ${\rm Time_{total}}$\\
\hline
Standard & 100& 6.0 & 590  \\
PGD-AT &120& 37 & 4440  \\
PGD-AT$^{\dag}$ & - & - & 4428  \\
TARDES & 120& 53 & 6360  \\
\hline
Free-AT$^{\dag}$ & 10 & 119 &  1188\\
FGSM-GA$^{\dag}$ & 30& 68 & 2052  \\
ATAS$^{\dag}$ & 30 & 36 & 1080  \\
\hline
vanilla AFT & 20 & 37 &  1330 \\
RiFT & 10 &  25 &  4690 \\
AFKD & 20 & 40 & 1390  \\
ARREST & 20 & 36  & 1310   \\
AFD & 20 & 46 & 1510  \\
\hline
\end{tabular}
\end{center}
\vskip -0.1in
\caption{Training time (second) of various methods on ResNet18 on CIFAR10. We show the time for one epoch (${\rm Time_{one}}$) and the expected values of the total time (${\rm Time_{total}}$). `Standard' denotes the natural pre-training.}
\label{time}
\end{table}

\section{Limitation and futural work}
\label{lim}
Subject to computation resources, we haven't conducted experiments on ImageNet-1k. We will explore effective adversarial fine-tuning approaches on large-resolution datasets in the future. Besides, we observe that the increasing gap of features also appears in adversarial training. It remains a lot for further studies. 

\section{Conclusion}

This paper uncovers a surprising increasing trend in the gap of features between natural and adversarial samples in AFT methods, and further investigates it from the perspective of features. We model features as the intrinsic features and confused features, in which the latter is defined as the specific latent features leading to the feature gap. Then we propose Adversarial Fine-tuning via Disentanglement (AFD) to bridge the feature gap to enhance robustness. We design the feature disentangler to explicitly separate out the confused features from features of adversarial samples. Besides, the disentangled features are aligned with the natural features in the pre-trained model. Experiments demonstrate that AFD effectively mitigates the feature gap and achieves a good trade-off between generalization and robustness.

\section{Acknowledgments}
This work was supported in part by the National Natural Science Foundation of China under Grants U22A2096, 62441601, and 62306227, in part by the Shaanxi Province Core Technology Research and Development Project under grant 2024QY2-GJHX-11, in part by the Fundamental Research Funds for the Central Universities under Grants QTZX23042 and ZYTS24142.



\bibliography{aaai25}

\clearpage

\appendix

\section{Further Discussion}

\emph{\textbf{Question 1:} What are the main differences of the motivation of this paper with the work to explore the existence of robust and non-robust features} \citep{ilyas2019adversarial}\emph{?}

These papers have different assumptions and definitions. (1) For \citet{ilyas2019adversarial}, they assume the features of adversarial samples are composed of robust and non-robust features. Robust features facilitate robustness as well as generalization (i.e., natural accuracy), and non-robust features benefit generalization but damage robustness. (2) In our work, considering the inconsistency of features leads to the vulnerability of DNN, we assume the ideal robust model should have a consistent understanding of natural and adversarial samples, thus extracting similar features from them. Besides, we divide the features into intrinsic and confused features. Intrinsic features contain clean representation. Confused features are induced by adversarial noise and result in a feature gap between natural and adversarial examples, thus causing wrong predictions. The definitions of features are different from \citet{ilyas2019adversarial}.

\emph{\textbf{Question 2:} As shown in Figure~\ref{issue_illustration} (a), the vanilla AFT also has a low feature distance and its distance is less than the proposed method from epoch 2 to epoch 14. Does it mean that the vanilla AFT gains better robustness?}

It is because the vanilla AFT did not converge well in the early stages of training, allowing the corresponding adversarial noise to successfully attack with only a small feature distance. However, it doesn’t change the increasing trend of the feature gap. As shown in Figure~\ref{issue_illustration} (a) after epoch 14, the feature distance of the vanilla AFT grows rapidly and exceeds AFD. Besides, its robust accuracy is lower than AFD all the time (an accuracy disparity of 1.2\%-2.8\%). When we enlarge the epochs to 40, our AFD rapidly converges to a small value in the feature distance and others still keep a low-speed increasing trend.

\begin{table}[!htbp]

\begin{center}
\begin{tabular}{ccccc}
\hline
Model&Data&Clean\\
\hline
ResNet18 & CIFAR-10 &  94.60  \\
ResNet18 & CIFAR-100 & 76.55  \\
ResNet18 & Tiny-ImageNet & 64.81  \\
\hline
ResNet50 & CIFAR-10 &  94.30  \\
ResNet50 & CIFAR-100 & 77.20  \\
\hline
WideResNet-34-10 & CIFAR-10 &  95.53 \\
WideResNet-34-10 &  CIFAR-100 &  80.28 \\
WideResNet-34-10 & Tiny-ImageNet & 67.86   \\
\hline
\end{tabular}
\end{center}
\caption{The clean accuracy of the pre-trained models.}
\label{pretrain}
\end{table}

\section{Experiments of Pre-trained models}
\label{details}
We naturally pre-train all the models with a batch size of 128, training epochs of 100, and a weight decay of ${5.0\times10^{-4}}$. The learning rate starts at 0.1 and then decays by $\times$0.1 with transition epochs $\{75, 90\}$, following~\citet{TRADES}. For AT methods, we train all the models with a batch size of 128, training epochs of 120, and a weight decay of ${2.0\times10^{-4}}$. The learning rate starts at 0.1 and then decays by $\times$0.1 with transition epochs $\{75, 90, 100\}$, following~\citet{mart}.

We conduct pre-training experiments on multiple datasets and architectures. As shown in Table~\ref{pretrain}, all of the pre-trained models have much higher accuracies than the fine-tuned models. It indicates that it is reasonable for feature alignment to regard features of the natural samples in the pre-trained models as the ideal instrinsic features.

\begin{figure}[!htbp]
\centerline{\includegraphics[width=\columnwidth]{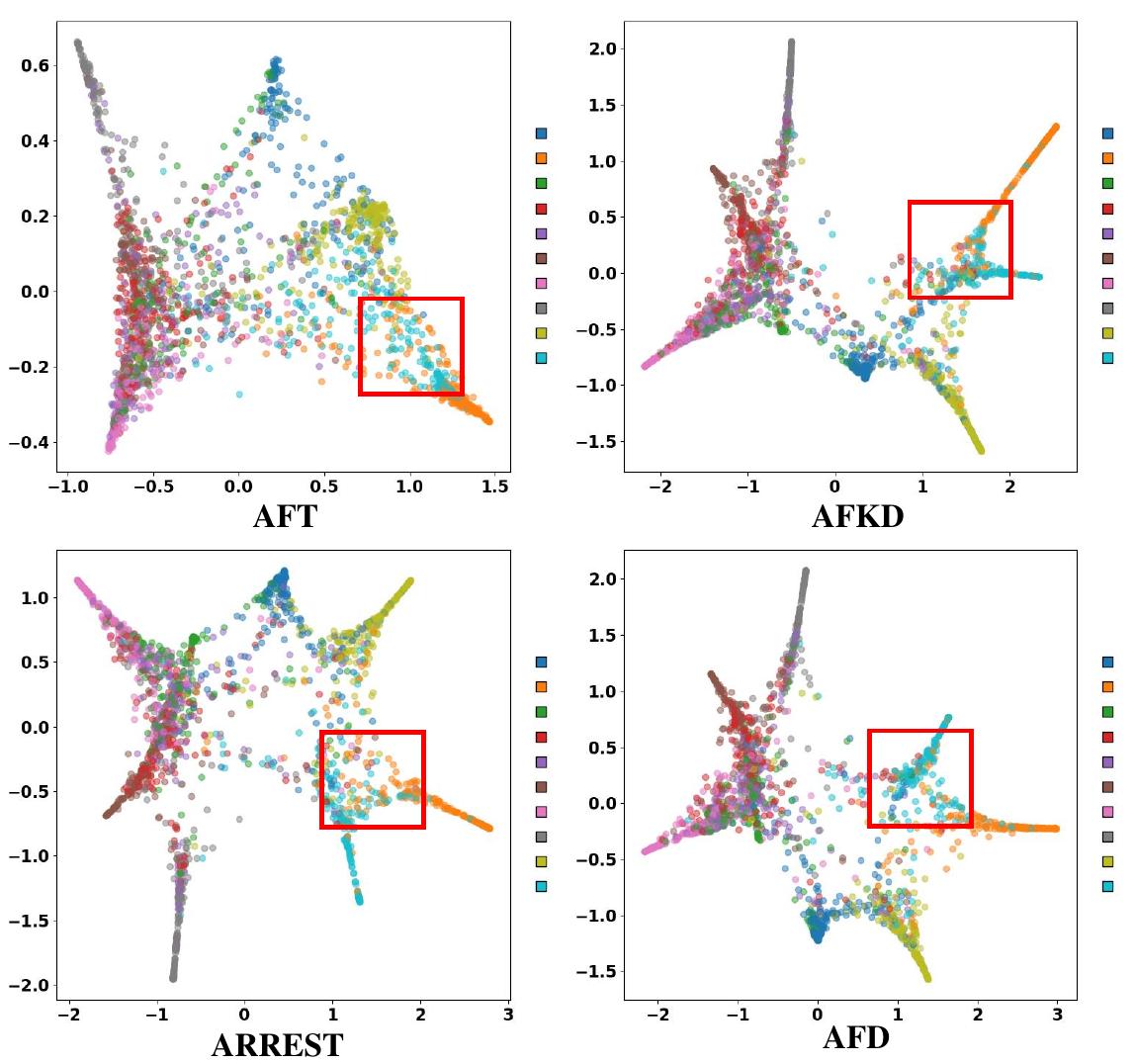}}
\caption{t-SNE visualization of different methods. Our AFD shows a clearer separation in the area of the red box.}
\label{t-sne}
\end{figure}

\section{Feature Visualization}
\label{visual}
 To further demonstrate the effectiveness of AFD, Figure~\ref{t-sne} visualizes the features of AFT methods using t-SNE~\citep{tsne} on ResNet18 on CIFAR-10. We color each point to match its ground-truth label and highlight focused areas with red boxes. It shows the disentangled features learned by AFD have a much clearer class boundary than those learned with other methods. This indicates that AFD successfully mitigates the confusion between features of distinct classes, leading to a more robust prediction.

\section{Supplement}
\label{supplement}
We have conducted experiments on a machine with three NVIDIA GeForce RTX 4090 GPUs. We use a single GPU for ResNet models and three GPUs for WideResNet models. We take three runs for each fine-tuning experiment, and the deviation is 0.43\%.

The gap between features of natural and adversarial samples can be observed in all the hidden layers. We select features of the \textbf{last} hidden layer to illustrate the gap since the $l_p$ norm of the feature gap in the last hidden layer is the largest. 

We have evaluated the feature gap in AFT by measuring the ${L}_1$, ${L}_2$, and ${L}_{\infty}$ distances between natural and adversarial features. The trends among different metrics are similar to Figure~1 (a).

The results in Figure 3 are obtained from the experiments on ResNet18 on CIFAR-10. The conclusions are similar among distinct datasets and models. The most important of the hyperparameters is $\gamma$, which can be searched in $\{10, 25, 30, 35\}$. Since AFD is insensitive to variations of hyperparameters, it is convenient to apply AFD on various datasets and architectures without searching hyperparameters too many times.

\end{document}